\documentclass[10pt,twocolumn,letterpaper]{article}

\usepackage[pagenumbers]{cvpr} % To force page numbers, e.g. for an arXiv version

\usepackage{pifont}
\usepackage{multirow}
\usepackage{algorithm}
\usepackage{algorithmic}
\newcommand{\cmark}{\ding{51}}

\newcommand{\xmark}{\textcolor{gray!70}{\ding{55}}} 
\definecolor{cvprblue}{rgb}{0.21,0.49,0.74}
\usepackage[pagebackref,breaklinks,colorlinks,citecolor=cvprblue]{hyperref}
\usepackage{graphicx}

\usepackage{tikz} % 用于绘制圆圈

\definecolor{pltblue}{RGB}{174, 199, 232}
\definecolor{pltorange}{RGB}{255, 229, 204}
\definecolor{pltgreen}{RGB}{204, 229, 204}
\definecolor{pltred}{RGB}{229, 204, 204}
\definecolor{pltpurple}{RGB}{239, 218, 230}

\definecolor{tabblue}{HTML}{1f77b4}
\definecolor{taborange}{HTML}{ff7f0e}
\definecolor{tabgreen}{HTML}{2ca02c}
\definecolor{tabred}{HTML}{d62728}
\definecolor{tabpurple}{HTML}{9467bd}
\definecolor{tabpink}{HTML}{ff0080}

\definecolor{cblue}{RGB}{173, 201, 233}
\definecolor{clblue}{RGB}{222, 234, 246}
\definecolor{corange}{RGB}{255, 152, 67}
\definecolor{lorgange}{RGB}{255, 221, 149}

\title{AeroGen: Enhancing Remote Sensing Object Detection with Diffusion-Driven Data Generation}

\author{Datao Tang$^1$\quad  Xiangyong Cao$^1$\thanks{Corresponding author.}\quad  Xuan Wu$^1$\quad  Jialin Li$^1$\\
Jing Yao$^4$\quad 
Xueru Bai$^2$\quad  Dongsheng Jiang$^3$\quad  Yin Li$^3$\quad  Deyu Meng$^1$ \\
$^1$Xi’an Jiaotong University \quad  $^2$Xidian University \\ $^3$Huawei Technologies Ltd \quad  $^4$Chinese Academy of Sciences\\
}

\begin{document}
\maketitle
\begin{abstract}
Remote sensing image object detection (RSIOD) aims to identify and locate specific objects within satellite or aerial imagery. However, there is a scarcity of labeled data in current RSIOD datasets, which significantly limits the performance of current detection algorithms. Although existing techniques, e.g., data augmentation and semi-supervised learning, can mitigate this scarcity issue to some extent, they are heavily dependent on high-quality labeled data and perform worse in rare object classes. To address this issue, this paper proposes a layout-controllable diffusion generative model (i.e. AeroGen) tailored for RSIOD. To our knowledge, AeroGen is the first model to simultaneously support horizontal and rotated bounding box condition generation, thus enabling the generation of high-quality synthetic images that meet specific layout and object category requirements. Additionally, we propose an end-to-end data augmentation framework that integrates a diversity-conditioned generator and a filtering mechanism to enhance both the diversity and quality of generated data. Experimental results demonstrate that the synthetic data produced by our method are of high quality and diversity. Furthermore, the synthetic RSIOD data can significantly improve the detection performance of existing RSIOD models, i.e., the mAP metrics on DIOR, DIOR-R, and HRSC datasets are improved by 3.7\%, 4.3\%, and 2.43\%, respectively. The code is available at \href{https://github.com/Sonettoo/AeroGen}{https://github.com/Sonettoo/AeroGen}.
\end{abstract}
%The effectiveness of our method was evaluated across three datasets, with experimental results demonstrating that the addition of synthetic data improved the mAP metrics by 3.7\%, 4.3\%, and 2.43\% on the DIOR, DIOR-R, and HRSC datasets, respectively.

\section{Introduction}

Object detection is a key technology for understanding and analyzing remote sensing images. It enables efficient processing of large-scale satellite data to extract and identify critical information, such as land cover changes \cite{zhang2020well}, urban development status \cite{li2020object}, and the impacts of natural disasters \cite{zheng2021building}. Through object detection, researchers can automatically extract terrestrial targets from complex remote sensing images, including buildings, vehicles, roads, bridges, farmlands, and forests. This information can be further applied in environmental monitoring, urban planning, land use analysis, and disaster emergency management.

With the rapid development of deep learning, supervised learning-based object detection algorithms have made significant progress in remote sensing image analysis \cite{zou2023object}. Although these algorithms can accurately locate and classify multiple objects in remote sensing images, they are heavily dependent on a large number of labelled training data. However, obtaining sufficient annotated data for remote sensing images is particularly challenging. Due to the presence of numerous and complex targets in remote sensing images, the manual annotation process is not only time-consuming and labour-intensive but also requires annotators to possess specialized knowledge, thus leading to high costs.

Although traditional data augmentation methods \cite{chlap2021review} (e.g., rotation and scaling) and enhancement techniques suitable for object detection (e.g., image mirror \cite{kisantal2019augmentation}, object-centric cropping \cite{liu2016ssd}, and copy-paste \cite{dwibedi2017cut}) can increase data diversity to some extent, they do not address the fundamental issue of insufficient data. The emergence of generative models \cite{nichol2021improved,ho2020denoising} provides a new solution to this problem. Currently, in the field of natural images, numerous high-performance generative models \cite{rombach2022high,podell2023sdxl} have been developed, capable of generating high-quality images from text conditions and also achieving significant progress in layout control. For remote sensing images, the application of generative models is usually combined with specific tasks, such as change detection  \cite{zheng2024changen2}, semantic segmentation \cite{toker2024satsynth} and road extraction \cite{tang2024crs}. These studies have been highly successful in utilizing data obtained from generative models to augment real-world datasets, thereby enhancing the performance of target models in downstream tasks. Therefore, utilizing generative diffusion models to fit the distribution of existing datasets and generate new samples to enhance the diversity and richness of remote sensing datasets is a feasible solution.

% Although traditional data augmentation methods  (such as rotation and scaling) alleviate the problem of insufficient data to some extent, they do not fundamentally address the core challenge of data deficiency.

In this paper, we focus on the remote sensing image object detection (RSIOD) task and construct a layout generation model (i.e., AeroGen) specifically designed for this task. The proposed AeroGen model allows for the specification of layout prior conditions with horizontal and rotated bounding boxes, enabling the generation of high-quality remote sensing images that meet specified conditions, thus filling a gap in the research field of RSIOD. Based on the AeroGen model, we further propose a conditional generation-based end-to-end data augmentation framework. Unlike pipeline-style data augmentation schemes in the natural image domain~\cite{zhao2023x}, our proposed pipeline is implemented by directly synthesizing RSIOD data through conditional generative models, thus eliminating the need for additional instance-pasting procedures. By introducing a diversity-conditioned generator and generation quality evaluation, we further enhance the diversity and quality of the generated images, thereby achieving end-to-end data augmentation for downstream object detection tasks. Moreover, we also design a novel filtering mechanism in this data augmentation pipeline to select high-quality synthetic training images, thus further boosting the performance.

In summary, the contributions of our work are threefold:
\begin{itemize}
\item We propose a layout-controllable diffusion model (i.e., AeroGen) specifically designed for remote sensing images. This model can generate high-quality RSIOD training datasets that conform to specified categories and spatial positions. To our knowledge, AeroGen is the first generative model to support layout conditional control for both horizontal and rotated bounding boxes. 

\item We design a novel end-to-end data augmentation framework that integrates the proposed AeroGen generative model with a layout condition generator as well as an image filter. This framework can produce synthetic RSIOD training datasets with high diversity and quality.

\item Experimental results show that the synthetic data can improve the performance of current RSIOD models, with improvements in mAP metrics by 3.7\%, 4.3\%, and 2.43\% on the DIOR, DIOR-R, and HRSC datasets, respectively. Notably, the performance in some rare object classes also significantly improves, e.g., achieving improvements of 17.8\%, 14.7\%, and 12.6\% in the GF, DAM, and APO categories, respectively.

\end{itemize}
\label{sec:intro}

% To insert a figure: \input{figs/template}
% Or table: \input{tables/template}
\section{Related Work}

\subsection{Diffusion Models}
Diffusion models \cite{ho2020denoising,nichol2021improved,rombach2022high}, known for their stable training process and excellent generative capabilities, are gradually replacing Generative Adversarial Networks (GANs) \cite{reed2016generative,zhang2017stackgan} as the dominant model in generative tasks. Text-guided diffusion models can produce realistic images, but the concise nature of text descriptions often makes it challenging to provide precise guidance for image generation, thereby limiting personalized generation capabilities. To address this issue, more researchers have introduced additional control conditions beyond text guidance, significantly expanding the application scope of diffusion models. These applications include layout guidance \cite{li2023gligen,zhou2024migc,wang2024instancediffusion}, style transfer \cite{wang2023stylediffusion,chung2024style}, image denoising and super-resolution \cite{chung2022mr}, and video generation \cite{ho2022imagen}, showcasing the enormous potential of diffusion models in various complex tasks. Among these models, LDM \cite{rombach2022high} restricts the diffusion process to a low-dimensional latent space, which not only preserves the high quality of the generated images but also significantly reduces computational complexity, serving as the foundation for numerous generative studies.

\subsection{Task-Oriented Data Generation}
The use of generative models to synthesize training data to assist in tasks like object detection \cite{zhu2024odgen, li2024simple}, semantic segmentation \cite{toker2024satsynth} and instance segmentation \cite{zhao2023x} has garnered significant attention from researchers. Generative models not only produce artistic natural images but also quickly adapt to specific industry scenarios such as remote sensing, medical, and industrial fields through techniques like fine-tuning. For instance, A Graikos et al. \cite{graikos2024learned} proposed representation-guided models that can generate embeddings rich in semantic and visual information through self-supervised learning (SSL), which guides the diffusion model to generate images. This approach reduces the difficulty of obtaining high-precision annotated data in specialized fields like histopathology and satellite imagery. In the area of remote sensing image generation, SatSynth \cite{toker2024satsynth} uses diffusion models to jointly learn the distribution of remote sensing images and their corresponding semantic segmentation labels. By generating semantically informed remote sensing images through joint sampling, it improves the performance of downstream segmentation tasks. Li Pang et al. \cite{pang2024hsigene} proposed a two-stage hyperspectral image (HSI) super-resolution framework that generates large amounts of realistic hyperspectral data for tasks like denoising and super-resolution. Moreover, models, e.g. CRS-Diff \cite{tang2024crs} and DiffusionSat \cite{khanna2023diffusionsat}, designed for optical remote sensing image generation, handle multiple types of conditional inputs, applying synthetic data to specific tasks such as road extraction. However, no existing research has specifically explored image generation methods for remote sensing image object detection (RSIOD) tasks. To fill this gap, we first propose a layout-controllable generative model that supports both rotated and horizontal bounding boxes, capable of synthesizing high-precision remote sensing images.

\subsection{Generative Data Augmentation}
To effectively apply synthetic data to downstream tasks, most existing methods directly combine synthetic data with real data for training. However, some studies (e.g., Auto Cherry-Picker \cite{chen2024auto}) have improved data quality by filtering synthetic data, thus better enhancing the performance of downstream tasks. For example, X-Paste \cite{zhao2023x} proposed a pipeline method that uses a copy-paste strategy to synthesize images, combined with a CLIP-based filtering mechanism, to further improve instance segmentation performance. A more comprehensive review of this issue can be found in DriverGen \cite{fan2024divergen}, which analyzes the application of synthetic data from the perspective of data distribution. It combines a copy-paste strategy to construct a multi-layer pipeline that enhances diversity and achieves significant results on the Lvis dataset \cite{gupta2019lvis}.

The most closely related approach to our work is ODGEN \cite{zhu2024odgen}, which employs an object-wise generation strategy to produce consistent data for multi-object scenes, addressing the domain gap and concept bleeding issues in image generation. In contrast, our work focuses on object detection in remote sensing images, utilizing a conditional generative model to directly synthesize data, thereby avoiding the additional instance-pasting process. Furthermore, we introduce a novel diversity-conditioned generator, combined with a filtering mechanism that accounts for both diversity and generation quality, to further enhance the diversity and quality of generated images. Through this approach, we achieve end-to-end data augmentation, significantly improving the performance of downstream tasks.
\label{sec:related}

\section{AeroGen}
\begin{figure*}
	\centering
	\includegraphics[width=0.95\linewidth]{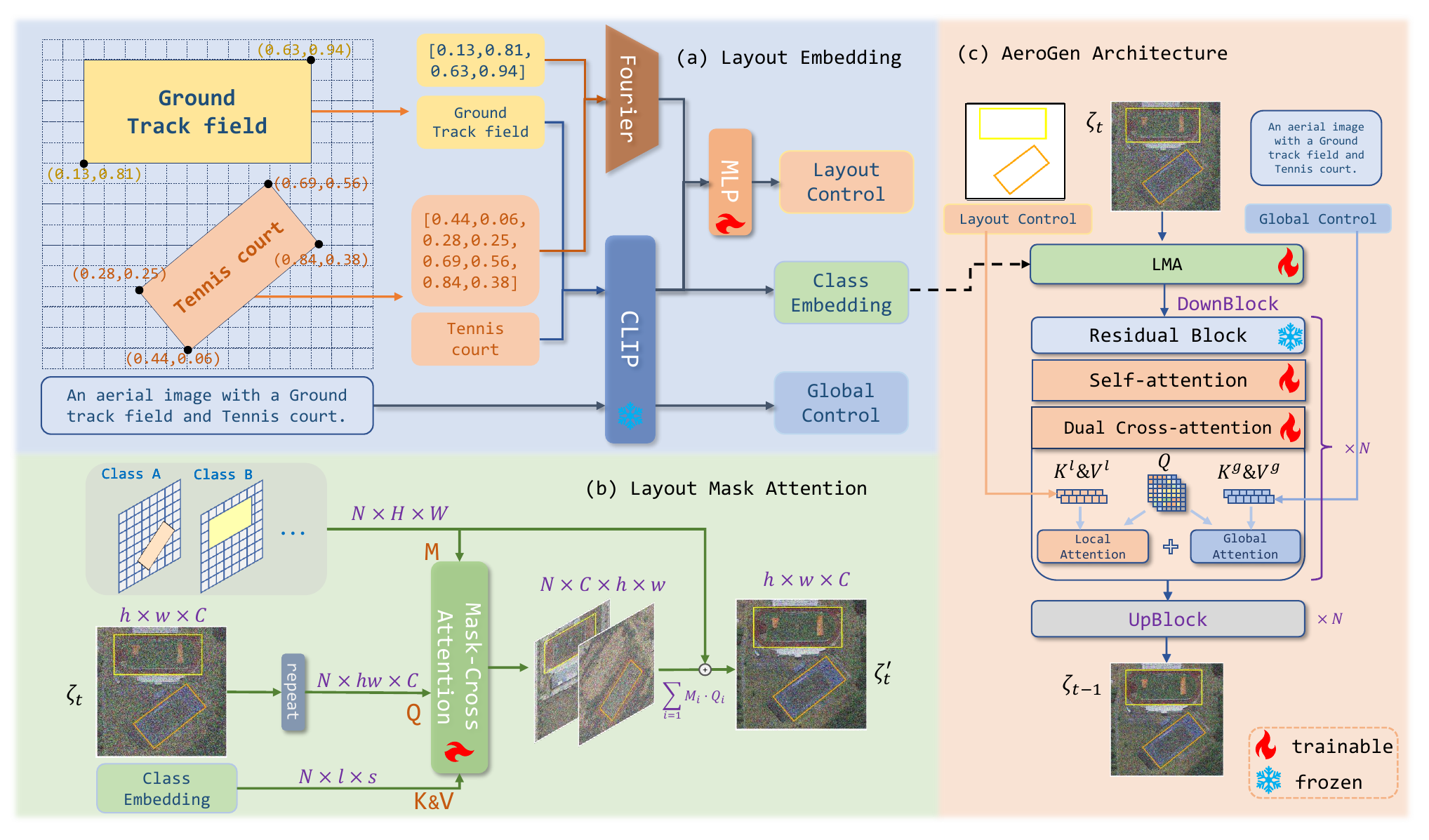}
	\caption{AeroGen's overall architecture. (a) The layout embedding module combines bounding box coordinates with vectorized semantic information using Fourier and MLP layers. This encodes layout information to facilitate control, with the prompt description processed by a CLIP text encoder for global conditional guidance. (b) The injection of layout information at the noise level is demonstrated, where a local mask governs the injection position of the layout information, allowing for finer layout control. (c) The overall architecture and training process of AeroGen. At each timestep, the image being denoised first passes through a layout information injection module, which enhances layout conditional guidance. }
\label{fig:model_1}
\end{figure*}

In this section, we introduce AeroGen, a layout-conditional diffusion model for enhancing remote sensing image data. The model consists of two key components: (a) a remote sensing image layout generation model in \cref{method3.1} that allows users to generate high-quality RS images based on predefined layout conditions, such as horizontal and rotated boxes; (b) a generation pipeline in \cref{3.2} that combines a diffusion-model-based diversity-conditional generator, which produces diverse layouts aligned with physical conditions, with a data-filtering mechanism to balance the diversity and quality of synthetic data, improving the utility of the generated dataset.

\subsection{Layout-conditional Diffusion Model}
\label{method3.1}

The model weights, obtained through comprehensive fine-tuning on a remote sensing dataset based on LDM \cite{rombach2022high,tang2024crs}, are adopted for RS study. In the original text-to-image diffusion model, the conditional position information is combined with the text control condition, and layout-based remote sensing image generation is achieved by establishing a unified position information encoding along with a corresponding dual cross-attention network, as shown in \cref{fig:model_1}. Building on the latest research advances, combined with the regional layout mask-attention strategy, control accuracy is improved, particularly for small target regions.

\textbf{Layout Embedding}. As shown in \cref{fig:model_1}(a), each object's bounding box or rotational bounding box is uniformly represented as a list of eight coordinates, i.e., $\mathbf{x} = [x_1, y_1, \dots, x_4, y_4]$, ensuring a consistent representation between horizontal and rotated bounding boxes. Building on this, Fourier \cite{mildenhall2021nerf} encoding is employed to convert these positional coordinates into a frequency domain vector representation, similar to GLIGEN \cite{li2023gligen}. We use a frozen CLIP text encoder \cite{radford2021learning} to obtain fixed codes $\mathbf{c}$ for different categories, which serve as layout condition inputs. The Fourier-encoded coordinates are then fused with the category encodings using an additional linear layer to produce the layout control input:
\begin{equation}
\mathbf{h} = \text{Linear}([\gamma(\mathbf{x}); \mathbf{c}]),
\end{equation}
where $[\gamma(\mathbf{x}); \mathbf{c}]$ denotes the concatenation of Fourier-coded coordinates and category codes, and $\text{Linear}(\cdot)$ represents the linear transformation layer. In this manner, spatial location and category information are effectively combined as layout control tokens.

\textbf{Layout Mask Attention}.
In addition to traditional token-based control, recent studies indicate that direct semantic embedding based on feature maps is also an effective method for layout guidance. In the denoising process of a diffusion model, the injection of conditional information is gradual, enabling local attribute editing at the noise level. To this end, conditionally encoded noise region steering is employed and combined with a cropping step for improved layout precision. As shown in \cref{fig:model_1}(b), each bounding box is first transformed into a 0/1 mask \( M \), and category attributes are obtained through CLIP encoding. During each denoising step, the mask attention network provides additional layout guidance. The process is expressed as follows: for each denoised image \( Q \) and category encoding \( K, V \), the mask \( M \) is used for attention computation according to the following equation:
\[
    \mathcal{Q} = \sum_{i=1}^{n} M_i \cdot \text{softmax}\left( \frac{Q K_i^\top}{\sqrt{d_k}} + \mathcal{M}_i \right)V_i,
\]
where \( M \) represents the corresponding bounding box mask, and \( \mathcal{M}_i \) derived from \( M_i \) as the attention mask. This method enables precise manipulation of local noise characteristics during the diffusion generation process, offering finer control over the image layout.

\textbf{AeroGen Architecture}.
In AeroGen, the text prompt serves as a global condition and is integrated with layout control tokens via a dual cross-attention mechanism. The output is computed as:
\begin{equation}
    \text{Out} = \Psi(Q, K^{\mathit{g}}, V^{\mathit{g}}) + \lambda \cdot \Psi(Q, K^{\mathit{l}}, V^{\mathit{l}}),
\end{equation}
where \(\Psi\) represents the cross-attention mechanism. \(K^{\mathit{g}}\) and \(V^{\mathit{g}}\) are the keys and values of the global text condition, while \(K^{\mathit{l}}\) and \(V^{\mathit{l}}\) are the layout control tokens. \(\lambda\) balances the influence of global and layout conditions.

The overall loss function for AeroGen combines both the global text condition and layout control, defined as:
\begin{equation}
    \mathcal{L} = \mathbb{E} \left[ \left\| \epsilon - \epsilon_{\theta}(\mathbf{x}_t, t, \mathbf{c}^{\mathit{g}}, \mathbf{c}^{\mathit{l}}) \right\|^2 \right],
\end{equation}
where \(\mathbf{x}_t\) represents the noisy image at time step \(t\), \(\mathbf{c}^{\mathit{g}}\) is the global text condition, and \(\mathbf{c}^{\mathit{l}}\) is the layout control. 

\begin{figure*}
	\centering
	\includegraphics[width=1\linewidth]{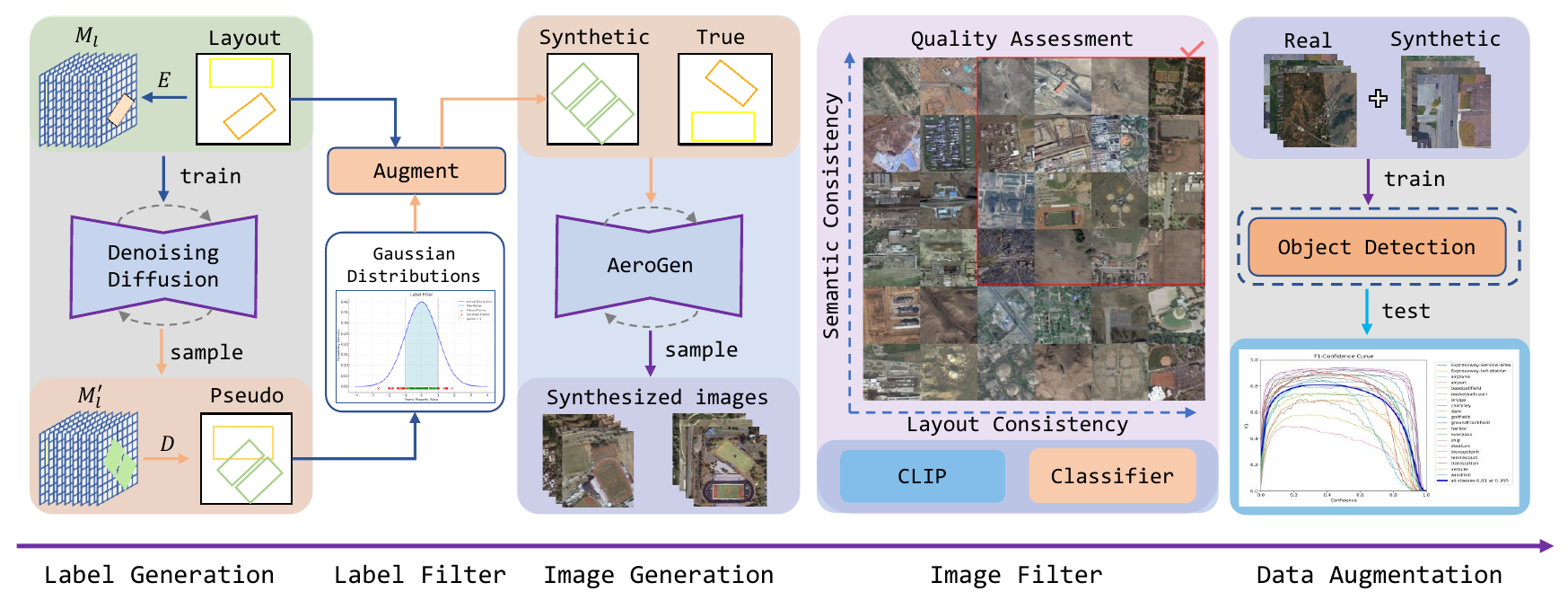}
	\caption{Overview of the pipeline based on AeroGen. By fitting the conditional distribution using a diffusion model, we expand a diverse set of layout conditions and combine them with AeroGen to generate synthetic data. Additionally, we introduce two filters to eliminate low-quality synthetic conditions and images, further ensuring the semantic consistency and layout consistency of the synthetic images. Finally, we incorporate synthetic images alongside real images in the training set to improve the performance of downstream tasks.}
\label{fig:pipeline}
\end{figure*}

\subsection{Generative Pipeline}
\label{3.2}
The layout generative pipeline, as illustrated in \cref{fig:pipeline}, is divided into five stages: label generation, label filter, image generation, image filter, and data augmentation. Each generation step is followed by a corresponding screening step to ensure synthesis quality.

\textbf{Label Generation}. 
Inspired by recent cutting-edge research \cite{toker2024satsynth}, we adopt a denoising diffusion probabilistic model (DDPM \cite{ho2020denoising}) to learn the conditional distribution and directly sample from it to obtain layout labels, thereby avoiding conflicts in layout conditions that may arise from random synthesis approaches. The specific method is illustrated in \cref{fig:pipeline}, where a labelling matrix \( \mathbf{M}_L \) is first constructed. This matrix contains all categories of conditions with dimensions \( H \times W \times N \), where \( H \) and \( W \) represent the height and width of the images, respectively, and \( N \) denotes the number of condition categories. For each condition corresponding to the target frame of the image, the value within the target frame region is set to 1, while the values in the remaining regions are set to -1. This process is formally represented as:
\begin{equation}
\mathbf{M}_L(i,j,k) = 
\begin{cases} 
    1, & \text{if} \ (i,j) \in \mathcal{R}_k, \\\\
    -1, & \text{if} \ (i,j) \notin \mathcal{R}_k, 
\end{cases}
\end{equation}
where \( i \in \{1, \dots, H\} \), \( j \in \{1, \dots, W\} \), and \( k \in \{1, \dots, N\} \), with \( \mathcal{R}_k \) denoting the target area for the \( k \)-th category. Next, this conditional distribution is fitted using a DDPM-based generator \( G_\theta \). The loss function is based on the mean square error (MSE):

\begin{equation}
\mathcal{L} = \mathbb{E} \left[ \|\epsilon - \epsilon_\theta(\mathbf{M}_L^{(t)}, t)\|^2 \right],
\end{equation}
where \( \mathbf{M}_L^{(0)} \) represents the original layout matrix, \( \mathbf{M}_L^{(t)} \) represents the noise matrix at the \( t \)-th time step, and \( \epsilon_\theta(\mathbf{M}_L^{(t)}, t) \) denotes the model's predicted noise at step \( t \).

\textbf{Label Filter and Enhancement}.
The label data sampled from the generator may not always align with real-world intuition or effectively guide image generation. Therefore, we propose a normal distribution-based filtering mechanism to screen the generated bounding box information, ensuring that the data conform to the distribution characteristics of real labels. The label filter assumes that the attributes of the bounding boxes (e.g., area) follow a normal distribution \((\mathcal{N}(\mu_X, \sigma_X^2))\) and introduces the following constraint: \(\frac{(X - \mu_X)}{\sigma_X} \leq \epsilon\), where \(\epsilon\) determines the filter's strictness, thereby ensuring that generated bounding boxes fall within a realistic and feasible range. Synthetic pseudo-labels and genuine a priori labels are filtered to form a comprehensive layout condition pool through additional enhancement strategies, including scaling, panning, rotating, and flipping.

\textbf{Image Generation}.
The synthetic bounding box labels are obtained based on the pool of layout conditions. The corresponding synthetic images are generated using the layout-guided diffusion model through the image generation process described in \cref{method3.1}. The model uses these bounding box labels to guide the generation, ensuring that the image content matches the generated layout conditions.

\textbf{Image Filter}.
Since the images generated by the diffusion model do not consistently meet high-quality or predefined layout requirements, a screening mechanism is implemented to evaluate both the quality of the generation and the consistency of the layout. The consistency of the semantic and layout is evaluated using the CLIP model \cite{liu2024remoteclip} and a ResNet101-based classifier \cite{he2016deep}. Synthetic images are then filtered by calculating their CLIP scores and minimum classification accuracies, which are compared against predefined thresholds to select the final filtered images.

\textbf{Data Augmentation}.
The synthetic data serves as a complementary dataset alongside the real dataset, and both are utilized as training data for downstream target detection model training.

\section{Experiments}

\begin{figure*}
	\centering
	\includegraphics[width=0.9\linewidth]{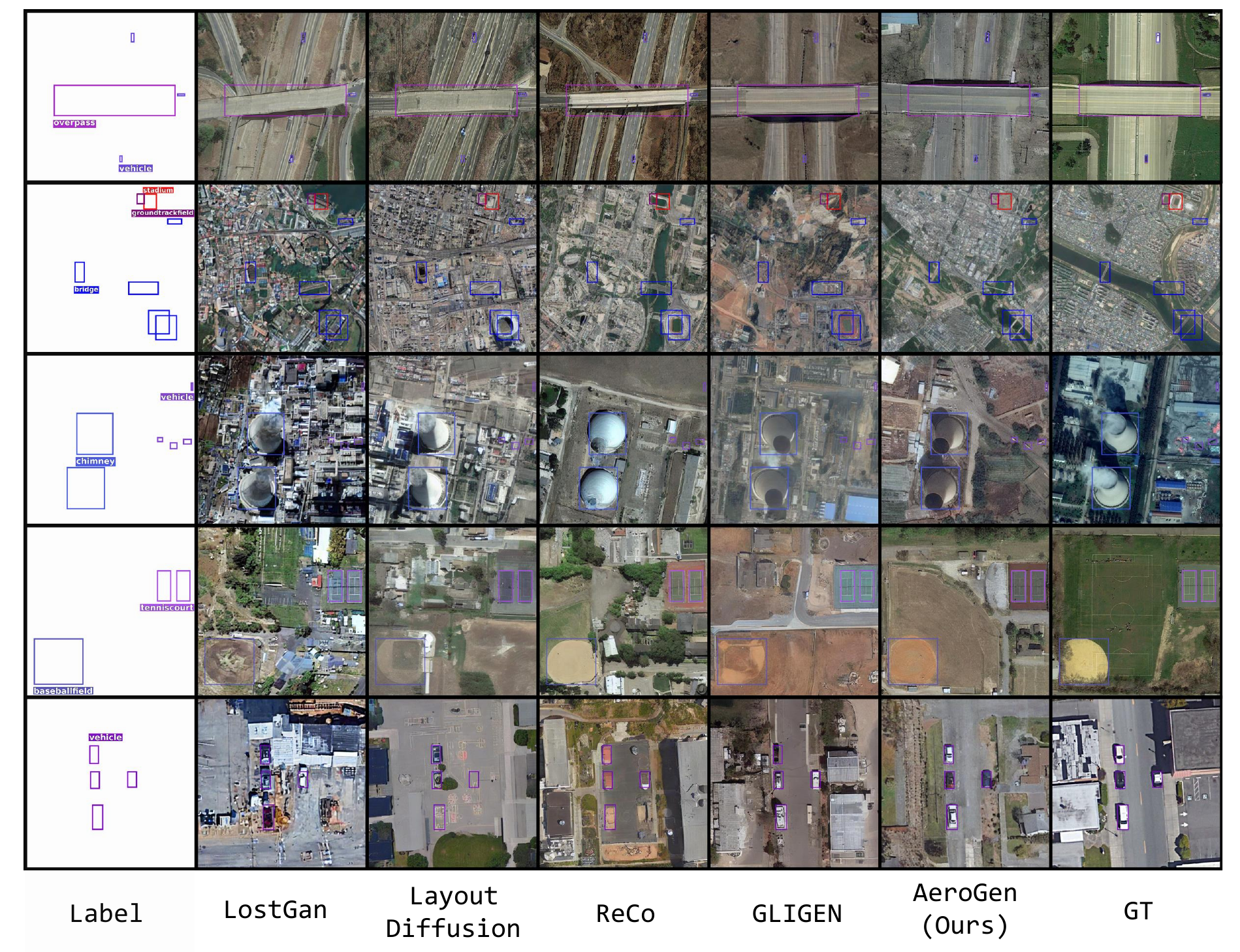}
	\caption{Visualization comparison of the generated image by different methods on the DIOR dataset. 
    % Compared to other methods, AeroGen produces superior quality and control and demonstrates exceptional performance in the layout control of small targets.
    }
\label{fig:display_2}
\end{figure*}

In this section, we conducted extensive experiments to verify the generative capabilities of AeroGen and its auxiliary data augmentation ability to support downstream RSIOD tasks. Specifically, we assessed the performance of our layout generation model AeroGen from both quantitative and qualitative perspectives. Subsequently, we performed data augmentation experiments on three datasets (i.e., DIOR, DIOR-R, and HRSC) to verify the effectiveness of synthetic data generated by our AeroGen model in improving the performance of downstream object detection tasks.

\subsection{Implementation Details}
\begin{table}
\centering
\small
\begin{tabular}{c|cccc}
\toprule
\textbf{Dataset} & \textbf{Modality} & \textbf{Images} & \textbf{Objects} & \textbf{Categories}  \\ \midrule

DIOR \cite{li2020object}  & HBB   & 23,463 &  192,518 & 20   \\

DIOR-R \cite{cheng2022anchor} & OBB  & 23,463 &  192,518 & 20   \\

HRSC \cite{liu2016ship}  & OBB   & 1,061  & 2,976   & 19   \\
% \cite{cheng2022anchor,li2020object,liu2016ship}
\bottomrule
\end{tabular}
\caption{Statistical information of the benchmark RSIOD datasets.}
\label{tab:dataset_info}
\end{table}

\textbf{Data Preparation.}
An overview of the three datasets is provided in \cref{tab:dataset_info}. Notably, the DIOR and DIOR-R datasets \cite{cheng2022anchor} share the same image data but differ in annotation format, with DIOR using bounding boxes and DIOR-R using rotated bounding boxes. HRSC \cite{liu2016ship} is a Remote Sensing dataset for ship detection, with image sizes ranging from 300 × 300 to 1500 × 900 pixels. It is divided into 436, 181, and 444 frames for training, evaluation, and testing, respectively. The DIOR/DIOR-R dataset is split into training, validation, and testing sets in a 1:1:2 ratio, with generative model training conducted exclusively on the training set.
% The three datasets are presented in \cref{tab:dataset_info}. Notably, DIOR and DIOR-R \cite{cheng2022anchor} share the same image data, though their annotation formats differ, with DIOR using bounding boxes and DIOR-R employing rotated bounding boxes.
% HRSC \cite{liu2016ship} is a remotely sensed dataset for ship detection, with image sizes ranging from 300 × 300 to 1500 × 900 pixels. The dataset is divided into 436, 181, and 444 frames for training, evaluation, and testing, respectively. The DIOR/DIOR-R dataset is divided into training, validation, and testing sets in a 1:1:2 ratio, with generative model training conducted exclusively on the training set.

\begin{table*}
\centering
\resizebox{0.9\textwidth}{!}{%
\begin{tabular}{@{\hspace{3em}}c@{\hspace{1.5em}}|@{\hspace{1.5em}}c@{\hspace{3em}}c@{\hspace{1.5em}}|@{\hspace{1.5em}}c@{\hspace{3em}}c@{\hspace{3em}}c@{\hspace{3em}}}

\toprule
\textbf{Method} & \textbf{Dataset} & \textbf{Modality} & \textbf{FID} $\downarrow$ & \textbf{CAS} $\uparrow$ & \textbf{YOLO Score} $\uparrow$ \\ \midrule

LostGAN \cite{sun2019image} & \multirow{6}{*}{DIOR \cite{li2020object}} & \multirow{6}{*}{HBB} & 57.10 & 46.02 & 14.3/27.3/15.2 \\ 

Layout Diffusion \cite{zheng2023layoutdiffusion} &  &  & 45.31 & 56.98 & 20.0/37.4/19.3 \\ 

ReCo \cite{yang2023reco} &  &  & 42.56 & 55.42 & 21.1/40.7/23.1 \\ 

GLIGEN \cite{li2023gligen} &  &  & 41.31 & 63.50 & 25.8/44.4/27.8 \\ 

\textbf{AeroGen (Ours)} &  &  & \textbf{38.57} & \textbf{76.84} & \textbf{29.8/54.2/31.6} \\ \midrule

GLIGEN \cite{li2023gligen}† & \multirow{2}{*}{DIOR-R \cite{cheng2022anchor}} & \multirow{2}{*}{OBB} & 48.43 & 58.89 & 24.6/41.6/25.1 \\ 

\textbf{AeroGen (Ours)} &  &  & \textbf{35.07} & \textbf{74.13} & \textbf{29.6/57.6/32.0} \\ \midrule

GLIGEN \cite{li2023gligen}† & \multirow{2}{*}{HRSC \cite{liu2016ship}} & \multirow{2}{*}{OBB} & 66.69 & 43.35 &23.4/44.7/26.3  \\ 

\textbf{AeroGen (Ours)} &  &  & \textbf{45.86} &\textbf{ 51.19}  &\textbf{27.1/51.0/27.6} \\ 

\bottomrule
\end{tabular}
}
\caption{Quantitative results of the generated images by different methods. For the Oriented Bounding Box (OBB) modality, we replicated the GLIGEN† method to account for the rotated bounding box. The best results are in \textbf{bold}.}
\vspace{-0.05in}
\label{tab:results_1}
\end{table*}

\textbf{Training Details.}
We trained our AeroGen separately on each dataset for 100 epochs. During training, we used the AdamW optimizer \cite{loshchilov2017decoupled} with a learning rate of 1e-5. Only the attention layers of UNet and the Layout Mask Attention (LMA) are updated, while the remaining weights are inherited from the fine-tuned LDM in RS data \cite{tang2024crs}.

\textbf{Evaluation Metrics.}
For the quantitative analysis of generated images, we used the FID score to evaluate the visual quality of the generated images and employed Classification Score (CAS) \cite{ravuri2019classification} and YOLO Score \cite{li2021image} to measure the layout consistency of the generated images. In the data augmentation experiments, we assessed object detection model performance based on mAP50 and mAP50-95 (mAP) metrics to evaluate their overall quality.

% \hspace{1em}
\subsection{Image Quality Results}
\textbf{Quantitative Evaluation.} We used a bounding box condition defined by four extreme coordinates and conducted both training and testing on the DIOR dataset. We compared AeroGen with state-of-the-art layout-to-image generation methods, including LostGAN \cite{sun2019image}, ReCo \cite{yang2023reco}, LayoutDiffusion\cite{zheng2023layoutdiffusion}, and GLIGEN \cite{li2023gligen}. The performance of these methods on three metrics is reported in \cref{tab:results_1}. To ensure fairness, we initialized all methods with identical SD weights and trained them on the DIOR dataset for the same number of epochs. Our method outperformed other methods across all the metrics.

Furthermore, we evaluated AeroGen and GLIGEN on the DIOR-R and HRSC datasets with rotated bounding boxes, where AeroGen consistently excelled. Notably, the original GliGen method does not support rotated bounding box conditions; therefore, we modified the layout encoding (as shown in \cref{fig:model_1} (a)) and retrained the model.

\textbf{Qualitative Evaluation.} \cref{fig:display_2} compares the results of AeroGen with those of other methods. AeroGen shows superior layout consistency and an enhanced capability for generating small objects. Besides, we present experimental results on natural images in the supplemental material.

\subsection{Data Augmentation Experiments}
We synthesize data on three RSIOD datasets for data augmentation. For the DIOR/DIOR-R datasets, we synthesized 10k, 20k, and 50k samples for the RSIOD task. For the HRSC dataset, we synthesise 2k, 4k, and 10k data in the same ratio. The training was performed using the OBB branch experimental setup of the unified YOLOv8 \cite{varghese2024yolov8} and Oriented R-CNN \cite{xie2021oriented} RSIOD models, and the model performance is verified on the corresponding test sets. The experimental results are shown in \cref{tab:gen_data_results}. The addition of synthetic data significantly improves performance on downstream tasks

% The target detection models trained on the joint dataset \( D \cup D' \) exhibit significantly higher detection accuracy for rare categories. 

\begin{table*}[h]
    \centering
    \scalebox{1}{
        \begin{tabular}{ccc}
            \begin{subtable}[t]{0.3\textwidth}
                \centering
                \begin{tabular}{c|cc}
                    \toprule[1pt]
                    \ \textbf{Gen Data} & \textbf{mAP} $\uparrow$ & \textbf{mAP50} $\uparrow$ \\
                    \midrule
                    0    & 54.22  & 72.69 \\
                    10k  & 55.62  & 74.79 \\
                    20k  & 56.78  & 76.31 \\
                    50k  & \textbf{57.92} & \textbf{77.10} \\
                    \bottomrule[1pt]
                \end{tabular}
                \caption{DIOR}
                \label{tab:DIOR_results}
            \end{subtable}
            &
            \begin{subtable}[t]{0.3\textwidth}
                \centering
                \begin{tabular}{c|cc}
                    \toprule[1pt]
                    \ \textbf{Gen Data} & \textbf{mAP} $\uparrow$ & \textbf{mAP50} $\uparrow$ \\
                    \midrule
                    0    & 37.39  & 60.21 \\
                    10k  & 39.81  & 62.39 \\
                    20k  & 41.12  & 63.33 \\
                    50k  & \textbf{41.69} & \textbf{64.12} \\
                    \bottomrule[1pt]
                \end{tabular}
                \caption{DIOR-R}
                \label{tab:DIOR_R_results}
            \end{subtable}
            &
            \begin{subtable}[t]{0.3\textwidth}
                \centering
                \begin{tabular}{c|cc}
                    \toprule[1pt]
                    \ \textbf{Gen Data} & \textbf{mAP} $\uparrow$ & \textbf{mAP50} $\uparrow$ \\
                    \midrule
                    0    & 63.49  & 90.28 \\
                    2k   & 64.12  & 91.79 \\
                    4k   & 64.78  & 92.31 \\
                    10k  & \textbf{65.92} & \textbf{93.10} \\
                    \bottomrule[1pt]
                \end{tabular}
                \caption{HRSC}
                \label{tab:HRSC_results}
            \end{subtable}

        \end{tabular}
    }
    \caption{The enhancement effects of various scales of synthetic data on downstream tasks across DIOR, DIOR-R, and HRSC datasets.}
    \label{tab:gen_data_results}
\end{table*}

\begin{figure}
	\centering
	\includegraphics[width=1\linewidth]{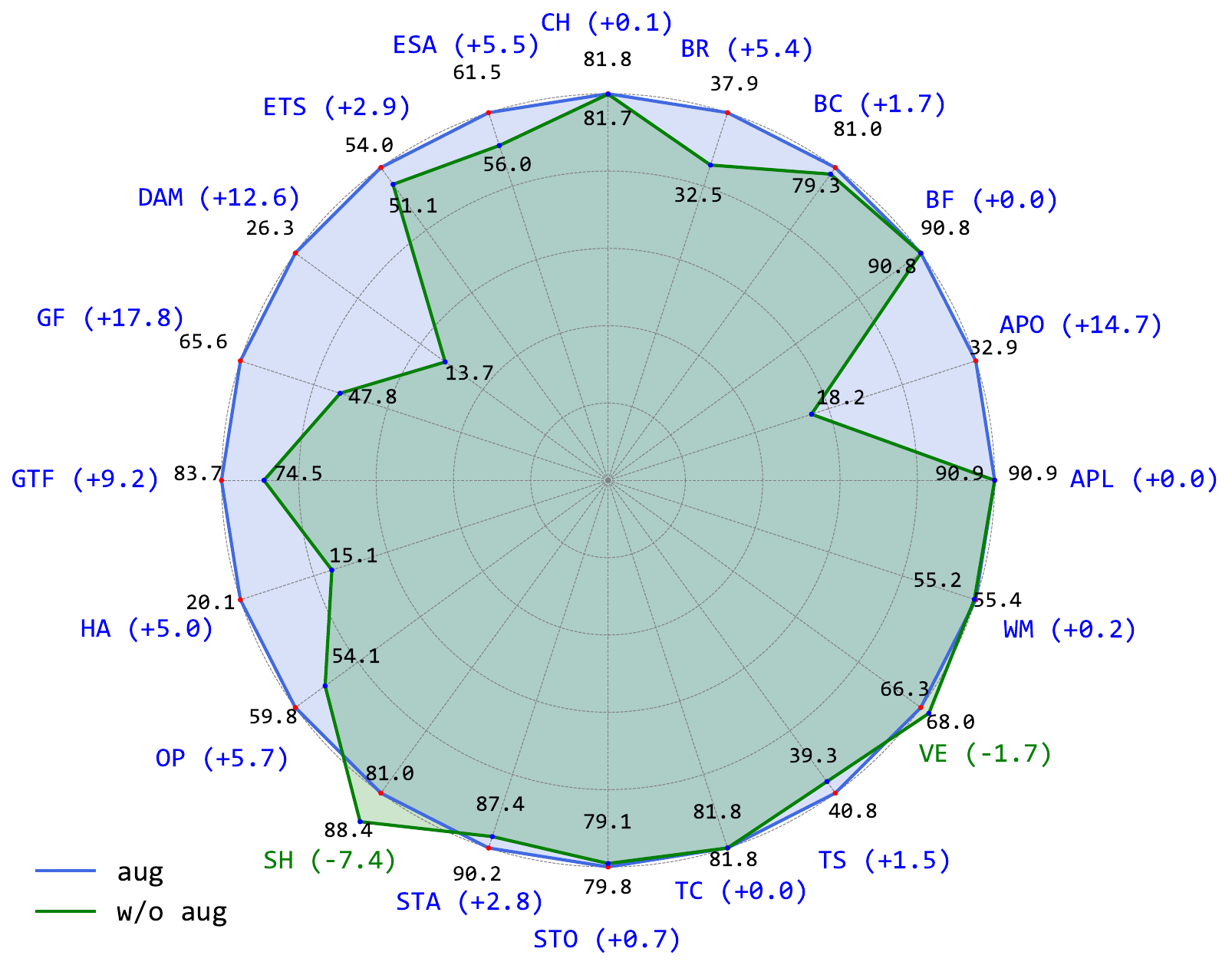}
	\caption{Comparison of mAP50 across each category on the DIOR-R dataset under the setting of data augmentation with 50k generation images (aug) and without augmentation (w/o aug). 
    %The \textcolor{blue}{blue} indicates rising categories, while the \textcolor{green}{green} indicates declining ones. 
 % The category abbreviations used in this study are: APL: airplane. APO: airport. BF: baseballfield. BC: basketballcourt. BR: bridge. CH: chimney. DAM: dam. ETS: expressway-toll-station. ESA: expressway-service-area. GF: golffield. GTF: groundtrackfield. HA: harbor. OP: overpass. SH: ship. STA: stadium. STO: storagetank. TC: tenniscourt. TS: trainstation. VE: vehicle. WM: windmill
 }
\label{fig:display_3}
\end{figure}

We visualize the mAP scores for different categories in detail, as shown in \cref{fig:display_3}. In most categories, results incorporating enhancements significantly outperform those without them, particularly in rarer categories, achieving improvements of 17.8\%, 14.7\%, and 12.6\% in the GF, DAM, and APO categories, respectively.

\begin{table}[h]
    \centering
	\scalebox{0.9}{
    \begin{tabular}{@{\hspace{1.5em}}c@{\hspace{1.5em}}|cc@{\hspace{1.5em}}}
        \toprule
        \textbf{Strategy} & \textbf{mAP} $\uparrow$ & \textbf{mAP50} $\uparrow$ \\
        \midrule
        Flip & 37.39 & 60.21 \\
        \midrule
        CopyPaste \cite{dwibedi2017cut}  & 38.25 & 61.79 \\
        AeroGen & 41.32 & 63.98 \\
        \midrule
        CopyPaste + Flip \cite{dwibedi2017cut} & 38.75 & 62.11 \\ 
        AeroGen + Flip & \textbf{41.69} & \textbf{64.12} \\
        \bottomrule
    \end{tabular}
    }
    \caption{Comparison of different augmentation strategies on the DIOR-R dataset.}
    \label{tab:strategy_performance}
\end{table}

\subsection{Ablation Study}
\textbf{Ablation of Enhanced Methods.} 
We compared synthetic data enhancement methods with traditional approaches, including the basic enhancement techniques of Flip and Copy-Paste \cite{dwibedi2017cut} for target detection tasks, as shown in \cref{tab:strategy_performance}. The target detection model trained on synthetic data performs significantly better than when trained with traditional methods, demonstrating the generative model’s effectiveness for data enhancement.

\textbf{Ablation of Different Modules.} 
We assessed the impact of different modules on image quality generated by AeroGen in \cref{tab:ablation}. The contribution of each module to the enhancement of image quality is evaluated by incorporating additional components into the original SD model. Results show that Layout Mask Attention (LMA) effectively captures global semantic information and preserves layout consistency, while adding Dual Cross Attention (DCA) further enhances performance, particularly in YOLO Score, indicating improved regional target generation. Overall, the model performs best when both LMA and DCA are used.

\begin{table}
	\centering
	\scalebox{0.95}{
		\begin{tabular}{cc|ccc}
			\toprule[1pt]
			\textbf{LMA} & \textbf{DCA} & \textbf{FID} $\downarrow$ & \textbf{CAS} $\uparrow$ & \textbf{YOLO Score} $\uparrow$ \\ 
			\midrule
			\xmark & \xmark & 82.11 & 18.48 & 1.3/3.9/1.1 \\ 
			\xmark & \cmark & 66.29 & 40.71 & 16.5/29.2/17.7 \\ 
			\cmark & \xmark & 61.50 & 50.11 & 25.3/46.5/27.2 \\ 
			\midrule
			\cmark & \cmark & \textbf{38.57} & \textbf{76.84} & \textbf{29.8/54.2/31.6} \\ 
			\bottomrule[1pt]
		\end{tabular}
	}
	\caption{Ablation study of different modules in the AeroGen model on the DIOR dataset.}
	\label{tab:ablation}
\end{table}

\begin{table}
	\centering
	\scalebox{0.85}{
		\begin{tabular}{ccc|cc|cc}
			\toprule[1pt]
			\multicolumn{3}{c|}{\textbf{Layout Diversity}} & \multicolumn{2}{c|}{\textbf{Image Quality}} & \multicolumn{2}{c}{\textbf{Metrics}} \\ 
			\midrule
			\rotatebox{60}{Synthesis} & \rotatebox{60}{Filter} & \rotatebox{60}{Augment} & \rotatebox{60}{Semantic} & \rotatebox{60}{Layout} & \rotatebox{60}{mAP $\uparrow$} & \rotatebox{60}{mAP50 $\uparrow$} \\ 
			\midrule
			\cmark & \cmark & \cmark & \cmark & \cmark & \textbf{41.69} & \textbf{64.12} \\ 
			\midrule
			\xmark & \xmark & \cmark & \cmark & \cmark & 41.31 & 63.47 \\ 
			\cmark & \xmark & \cmark & \cmark & \cmark & 40.92 & 62.41 \\ 
			\cmark & \cmark & \xmark & \cmark & \cmark & 39.62 & 61.32 \\ 
			\cmark & \cmark & \cmark & \xmark & \cmark & 40.27 & 62.13 \\ 
			\cmark & \cmark & \cmark & \cmark & \xmark & 37.05 & 60.03 \\ 
			\bottomrule[1pt]
		\end{tabular}
	}
	\caption{Ablation study results analyzing the impact of diverse generation and filtering strategies in the synthesis pipeline on layout labeling and image filtering consistency.}
	\label{tab:filter_strategy}
\end{table}

\textbf{Ablation of Augment Pipeline.}
We further analyze the filtering strategies and data augmentation techniques in the generation pipeline, including diverse generation strategies, filtering strategies for layout conditions, and filtering strategies for layout and semantic consistency of images. We use the synthetic data generated in various ways as enhancement data and conduct enhancement experiments on the DIOR-R datasets and the experimental results are shown in \cref{tab:filter_strategy}. As can be seen, each component in the generation pipeline contributes positively.

\label{sec:experiments}

\section{Conclusion}
This paper introduces AeroGen, a layout-controllable diffusion model designed to enhance remote sensing image datasets for target detection. The model comprises two primary components: a layout generation model that creates high-quality remote sensing images based on predefined layout conditions, and a data generation pipeline that incorporates a diversity of condition generators for the diffusion model. The pipeline employs a double filtering mechanism to exclude low-quality generation conditions and images, thereby ensuring the semantic and layout consistency of the generated images. By combining synthetic and real images in the training set, AeroGen significantly improves model performance in downstream tasks. This work highlights the potential of generative modeling in enhancing the datasets of remote sensing image processing tasks.
\label{sec:conclusion}

{
    \small
    \bibliographystyle{ieeenat_fullname}
    \bibliography{11_references}
}

% \input{12_appendix}

\end{document}